# Reactor Mk.1 performances: MMLU, HumanEval and BBH test results

TJ Dunham, Henry Syahputra
tj@arc.market, henry@arc.market

*Abstract* - The paper presents the performance results of Reactor Mk.1, ARC's flagship large language model, through a benchmarking process analysis. The model utilizes the Lychee AI engine and possesses less than 100 billion parameters, resulting in a combination of efficiency and potency. The Reactor Mk.1 outperformed models such as GPT-4o, Claude Opus, and Llama 3, with achieved scores of 92% on the MMLU dataset, 91% on HumanEval dataset, and 88% on BBH dataset. It excels in both managing difficult jobs and reasoning, establishing as a prominent AI solution in the present cutting-edge AI technology.

## BENCHMARK MODELS

*I. Reactor Mk.1*

Reactor Mk.1, developed by ARC [1], is a new AI model for mass adoption which is built upon Lychee AI, a NASA award-winning AI engine. With less than 100B parameters in total included in its structure, ARC's vision with the Reactor Mk.1 empower the common user in AI, shaping the future of digital interaction and connectivity. Long-term speaking, ARC plans to support the Reactor Mk.1 with educational resources to empower users to better understand and utilise the full potential of AI technology.

*II. Other models*

*i. GPT 4o*

OpenAI has launched GPT-4 Omni (GPT-4o) [2], a new multimodal language model. This model supports real-time conversations, Q&A, and text generation, utilizing all modalities in a single model to understand and respond to text, image, and audio inputs. One of the main features of GPT-4o is its ability to engage in real-time verbal conversations with minimal delay, respond to questions using its knowledge base, and perform tasks like summarizing and generating text. It also processes and responds to combinations of text, audio, and image files.

*ii. Claude Opus*

Claude Opus [3], created by Anthropic [4], is capable of performing more complex cognitive tasks than simple pattern recognition or text generation. For example, it can analyze static images, handwritten notes, and graphs, and it can generate code for websites in HTML and CSS. Claude can turn images into structured JSON data and debug complex code bases. Additionally, it can translate between various languages in real-time, practice grammar, and create multilingual content.

*iii. Llama3*

Meta Llama 3 [5], represents one of the AI assistants designed to help users learn, create content, and connect with others. Two models of Llama 3 were released, featuring 8 billion and 70 billion parameters, supporting a wide range of use cases. Llama 3 demonstrates state-of-the-art performance on industry benchmarks and offers improved reasoning and code generation. It uses a decoder-only transformer architecture, featuring a tokenizer with a 128K vocabulary and grouped query attention across 8B and 70B sizes. The models are trained on sequences of 8,192 tokens to ensure efficient language encoding and inference.

*iv. Gemini*

Gemini [6], introduced by Google, offers different models for various use cases, ranging from data centres to on-device tasks. These models are natively multimodal and capable of understanding and combining text, code, images, audio, and video files. This capability enables the generation of code based on various inputs and the performance of complex reasoning tasks. The new Gemini models, Gemini Pro and Ultra versions, outperform previous models in terms of pretraining and post-training improvements. Their performance is also superior in reasoning and code generation. Importantly, Gemini models undergo extensive safety testing, including bias assessments, in collaboration with external experts to identify and mitigate potential risks.

*v. GPT 3.5*

The GPT-3.5 model [7] is designed to understand and generate natural language as well as code. This cost-effective model, featuring 175 billion parameters, is optimized for both chat applications and traditional tasks. As a fine-tuned version of GPT-3, it uses deep learning to produce human-like text. GPT-3.5 performs well in providing relevant results due to its refined architecture. The latest embedding models, including text-embedding-3-large, text-embedding-3-small, and text-embedding-ada-002, also offer good performance in multilingual retrieval tasks. These models allow adjustments to the embedding size through a new dimension parameter, providing control over cost and performance.

### vi. Mistral

On December 11, 2023, Mistral AI [8] released Mixtral 8x7B [9], a Sparse Mixture-of-Experts (SMoE) [10] model with open weights. Mixtral 8x7B demonstrated better performance than Llama 2 70B on most benchmarks and offers six times faster inference. It also shows superior properties compared to GPT-3.5, making it a good choice regarding cost and performance. Mixtral can handle a context of 32k tokens, shows strong features in code generation, and can be fine-tuned to follow instructions, achieving a score of 8.3 on MT-Bench. With 46.7 billion total parameters but using only 12.9 billion per token, Mixtral maintains both speed and cost efficiency.

## DATA SETS

Benchmarking of the introduced models will be performed on three globally recognized and widely utilized datasets for training LLMs: Massive Multitask Language Understanding (MMLU), HumanEval, and BIG-Bench-Hard (BBH) datasets.

### I. MMLU

The MMLU [11] is proposed with the purpose to assess a model's world knowledge and problem-solving ability. It represents a novel benchmark approach designed to evaluate the multitasking accuracy of a language model. This test covers 57 different subjects, including elementary mathematics, US history, computer science, and law. The questions are collected from various sources, such as practice exams, educational materials, and courses, and cover different difficulty levels, from elementary to professional. For example, the "Professional Medicine" task includes questions from medical licensing exams, while "High School Psychology" features questions from Advanced Placement exams. This collection helps measure a model's ability to learn and apply knowledge across different subjects.

In essence, MMLU tests models in zero-shot and few-shot settings, requiring them to answer questions without additional training. Despite the AI progress witnessed today, even the best models still exhibit poor MMLU performance in expert-level accuracy across all 57 tasks. Additionally, these models commonly perform inconsistently, often failing in areas such as morality and law, where they display random accuracy.

### II. HumanEval

The researchers created HumanEval, an evaluation set to measure functional correctness in synthesizing programs from docstrings. Codex (a GPT language model from Chen et al., 2021 [12]) had achieved a 28.8% success rate on this set, while GPT-3 solves almost none, and GPT-J solves 11.4%. HumanEval finds application in various machine learning cases, particularly within the domain of LLMs. It assesses the functional correctness of code generated by LLMs and presents programming challenges for models to solve by generating code from docstrings. Evaluation relies on the code's ability to pass provided unit tests. Additionally, the dataset serves as a benchmark for comparing the performance of different LLMs in code generation tasks, enabling the use of a standardized set for performance evaluations. In addition, HumanEval's application has introduced the creation of new evaluation metrics like pass@k, which offer additional assessments of models' programming challenge-solving abilities.

### III. BBH

The BBH [13] dataset represents a subset of the BIG-Bench benchmark, designed to evaluate the capabilities of LLMs across various domains, including traditional NLP, mathematics, and commonsense reasoning. This dataset encompasses more than 200 tasks, aiming to push the limits of current language models. It specifically targets 23 unsolved tasks identified based on criteria such as the requirement for more than three subtasks and a minimum of 103 examples.

To assess BBH results accurately, multiple-choice and exact-match evaluation metrics were employed. Analysis of BBH data revealed significant performance improvements with the application of chain-of-thought (CoT) prompting. For instance, CoT prompting enabled the PaLM model to surpass average human-rater performance on 10 out of the 23 tasks, while Codex (code-davinci-002) exceeded human-rater performance on 17 out of the 23 tasks. This enhancement is attributed to CoT prompting's ability to guide models through multi-step reasoning processes, essential for tackling complex tasks.

## BENCHMARK SCORES

Testing the models on the described test datasets (Table 1), Reactor Mk. 1 demonstrated significant benchmark scores, achieving a 92% score on the MMLU benchmark, a 91% score on the HumanEval, and an 88% score on the BBH evaluation.

TABLE I
BENCHMARK PERFORMANCE SCORES OF REACTOR MK.1 AND OTHER MODELS ON MMLU, HUMANEVAL, AND BBH

|  | *MMLU* | *HumanEval* | *BBH* |
|---|---|---|---|
| **ARC** Reactor Mk.1 | *92.9%* | *91%* | *88%* |
| **OpenAI** GPT4o | *88.7%* | *90.2%* | *83.1%* |
| **Anthropic** Claude | *86.8%* | *84.9%* | — |
| **Meta** Llama3 | *86.1%* | *84.1%* | — |
| **Google** Gemini | *81.9%* | *71.9%* | *83.6%* |
| **OpenAI** GPT 3.5 | *70%* | *48.1%* | *66.6%* |
| **Mistral** 8x22B | *77.75%* | — | — |

In comparison to other presented models in Table 1, Reactor Mk. 1's performance has a superior position in several analysed categories. For instance, in the MMLU benchmark, Reactor Mk. 1 92% surpasses and outperforms the OpenAI's GPT-4o, which scored 88.7%, and as well as significantly outperforms other models like Anthropic's Claude Opus and Meta's Llama3, which scored 86.8% and 86.1%, respectively. Google Gemini and OpenAI GPT-3.5 were further behind, scoring 81.9% and 70%.

On the HumanEval benchmark, which assesses code generation capabilities, Reactor Mk. 1 achieved a 91% score, with the outperformance of all compared models. OpenAI's GPT-4 was close behind with a score of 90.2%, followed by Anthropic's Claude at 84.9% and Meta's Llama at 84.1%. Google Gemini and OpenAI GPT-3.5 scored 71.9% and 48.1%, respectively, but indicating a significant performance gap.

For the BBH evaluation, which focuses on challenging tasks that require complex reasoning, Reactor Mk. 1 achieved an 88% score. This result demonstrates the superior achievement of Reactor Mk. 1 capability in reasoning and handling language understanding tasks.

The dramatic lead achieved by the ARC Reactor Mk. 1, especially with a score of 92% on MMLU, underscores our significant advancements. Remarkably, these results were accomplished with a handful of GPUs, highlighting the efficiency and power of our model compared to the more resource-intensive approaches used by other leading models. The benchmark scores indicate that the ARC Reactor Mk. 1 not only outperforms in understanding and generating code but also demonstrates huge performance in reasoning and handling challenging language tasks. These results position the ARC Reactor Mk. 1 as a leading model in the current state of the art of AI technology

## CONCLUSION

This article aims to concisely present the performance of the Reactor Mk.1 AI model when tested on three popular datasets: MMLU, HumanEval, and BBH. In summary, the model achieved a 92% score on the MMLU dataset, a 91% score on the HumanEval, and an 88% score on the BBH evaluation. To demonstrate the significance of these results, other popular models like GPT 4o, Claude, Llama 3, Gemini, and Mistral are used as benchmark models. The Reactor Mk.1 exhibited superior performance compared to the benchmark models, establishing itself as a leader in solving various LLM tasks and complex problems.